\documentclass[runningheads]{llncs}
\usepackage[T1]{fontenc}
\usepackage{color, soul} 
\usepackage{amsfonts}
\usepackage{amssymb}
\usepackage{hyperref}
\hypersetup{hidelinks,
	colorlinks=true,
	allcolors=black,
	pdfstartview=Fit,
	breaklinks=true}

\usepackage{marvosym}
\usepackage{graphicx}
\usepackage{epstopdf}

\usepackage{multirow}
\usepackage{amsmath}
\usepackage{booktabs}
\usepackage{tabulary}
\usepackage{tabularx}
\usepackage{array}
\usepackage{ulem}
\usepackage{xcolor}

\makeatletter
\def\@author{ }
\makeatother

\makeatletter
\def\@institute{ }
\makeatother

\begin{document}
\title{EPPS: Advanced Polyp Segmentation via Edge Information Injection and Selective Feature Decoupling}
\titlerunning{Edge-Prioritized Polyp Segmentation}
%
\author{Mengqi Lei\inst{1} \and Xin Wang\inst{2}\textsuperscript{(\Letter)}}

\authorrunning{Lei et al.}

\institute{
China University of Geosciences, Wuhan 430074, China\\
\and
Baidu Inc, Beijing, China\\
\email{wangxin105@baidu.com}
}
%
\maketitle              

\begin{abstract}
Accurate segmentation of polyps in colonoscopy images is essential for early-stage diagnosis and management of colorectal cancer. Despite advancements in deep learning for polyp segmentation, enduring limitations persist. The edges of polyps are typically ambiguous, making them difficult to discern from the background, and the model performance is often compromised by the influence of irrelevant or unimportant features. To alleviate these challenges, we propose a novel model named Edge-Prioritized Polyp Segmentation (EPPS). Specifically, we incorporate an Edge Mapping Engine (EME) aimed at accurately extracting the edges of polyps. Subsequently, an Edge Information Injector (EII) is devised to augment the mask prediction by injecting the captured edge information into Decoder blocks. Furthermore, we introduce a component called Selective Feature Decoupler (SFD) to suppress the influence of noise and extraneous features on the model. Extensive experiments on 3 widely used polyp segmentation benchmarks demonstrate the superior performance of our method compared with other state-of-the-art approaches. The code is available at \url{https://github.com/Mengqi-Lei/Edge-Prioritized-Polyp-Segmentation}.

\end{abstract}

\section{Introduction}
 
\begin{figure}[t]
\includegraphics[width=1\linewidth]{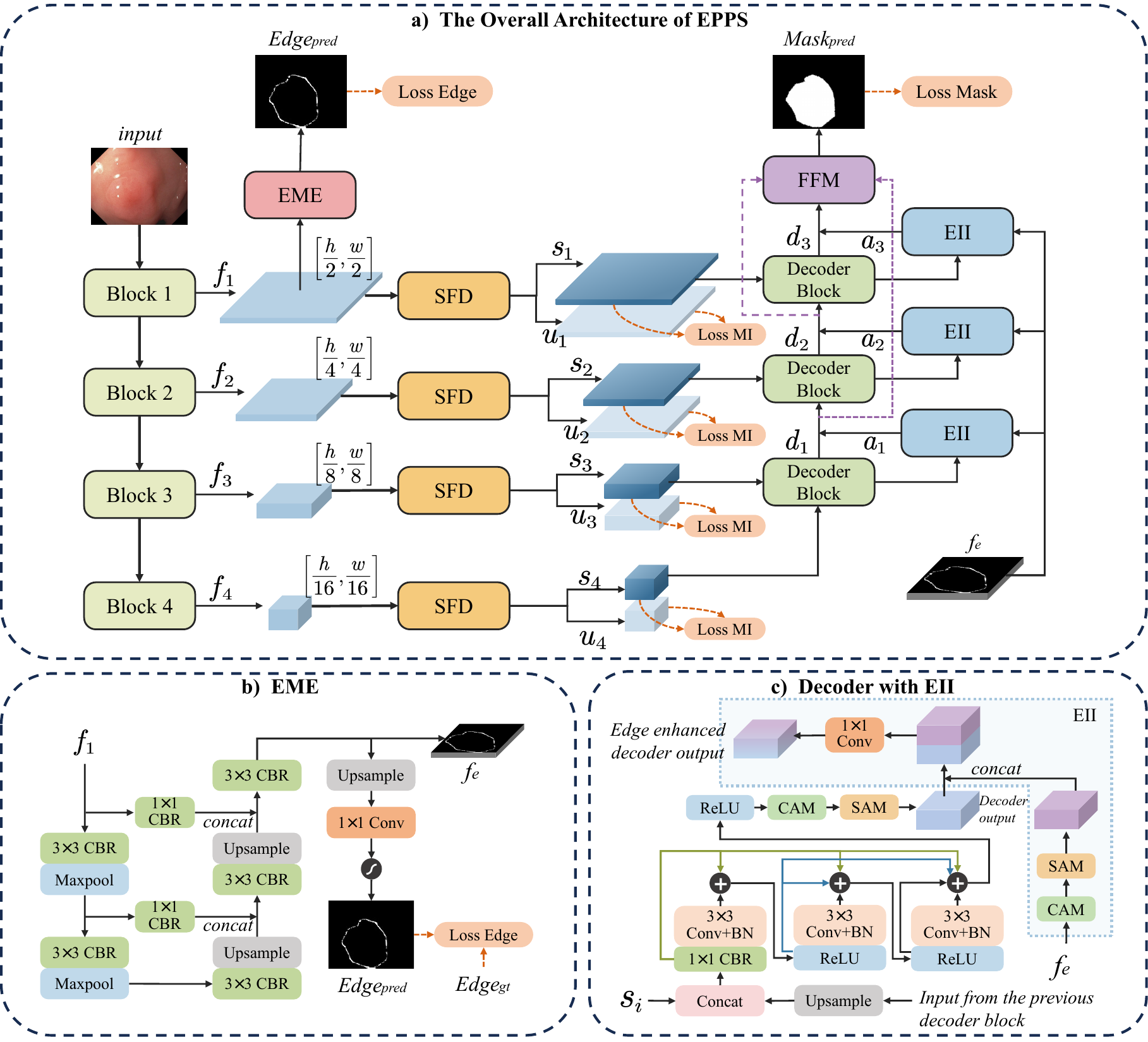}
\caption{a)The overall architecture of EPPS. b)The Edge Mapping Engine. c)The Decoder block with Edge Information Injector. In this figure, CBR represents Convolution, Batch Normalization and ReLU. CAM and SAM stand for Channel Attention Module and Spatial Attention Module, respectively.} \label{fig_major}
\end{figure}

Colorectal cancer (CRC) persists as a critical public health concern, contributing to elevated mortality rates worldwide \cite{first_ref}.
CRC often originates from adenomatous polyps, and a polyp usually takes ten to fifteen years to develop into cancer. 
Therefore, effective detection and removal of polyps before they become malignant can prevent the occurrence of CRC. 
Colonoscopy is considered the gold standard for detecting colorectal polyps. 
During colonoscopy, polyp segmentation is a fundamental task in computer-aided detection (CADe) of polyps, which is of significant importance in the clinical prevention of CRC \cite{third_ref,second_ref}.

In recent years, Encoder-Decoder based deep learning models have made significant progress in polyp segmentation \cite{systematic}, such as U-Net \cite{unet}, R2U-Net \cite{r-unet}, PraNet \cite{pranet}, HRENet \cite{hrenet}, TGANet \cite{tganet} and LDNet \cite{ldnet}. 
However, existing methods still face challenges in accurately locating polyps. On one hand, the edges of polyps are often difficult to clearly distinguish from the surrounding mucosa, which is a significant difficulty in polyp segmentation. This ambiguity is particularly evident when dealing with flat lesions or when the bowel preparation is insufficient. On the other hand, due to the complexity and variability of the background and different lighting conditions, models are easily disturbed by a large number of irrelevant or unimportant features, resulting in limited performance.

To alleviate these issues, we propose a method called Edge-Prioritized Polyp Segmentation (EPPS). Firstly, we designed an Edge Mapping Engine (EME) module to precisely extract edge information from the feature maps outputted by the Encoder. Specifically, EME utilizes a multi-level downsampling strategy to enhance edge features, followed by a special upsampling process to refine and accurately restore edge details, thereby achieving precise edge prediction.
Simultaneously, we introduce the Edge Information Injector (EII), which employs serial channel and spatial attention mechanisms \cite{cbam} to inject the edge information captured by EME into the Decoder's feature maps, guiding the prediction of the polyp mask.
Furthermore, to avoid the negative impact of ineffective features on model performance, we propose a Selective Feature Decoupler (SFD) module. It serves as a conduit connecting the Encoder and Decoder. 
By estimating and minimizing mutual information through a neural network, the feature map is decoupled into important and unimportant information, thus achieving filtration of ineffective information in the feature map.
Notably, our approach is entirely end-to-end, requiring no manual annotation of polyp boundaries, nor any phase-specific separate training.

Overall, our main contributions to precise polyp segmentation are as follows: 

(1) By introducing the Edge Mapping Engine (EME) and Edge Information Injector (EII) modules, we efficiently guide the model to distinguish the edges of polyps, thereby achieving more accurate polyp mask predictions. 

(2) The proposed Selective Feature Decoupler (SFD) can separate ineffective information from the feature map, filtering features during the transfer process from the Encoder to the Decoder.

(3) Our EPPS has achieved state-of-the-art performance on the Kvasir-SEG \cite{kvasir_seg}, CVC-ClinicDB \cite{cvc_clinicdb}, and Kvasir-Sessile \cite{kvasir_seg} datasets. Extensive experiments have validated the effectiveness and robustness of the modules we propose.

\section{Methodology}

\subsubsection{Overview.}
As shown in Fig.\ref{fig_major}(a), EPPS is a model based on the Encoder-Decoder architecture, primarily divided into four steps. (1) We use ResNet-50 \cite{resnet} as the Encoder, which extracts feature maps of different scales at its various stages to obtain feature maps \(f_1\) to \(f_4\) containing diverse semantic information. (2) The feature map \(f_1\) is fed into the Edge Mapping Engine (EME) to extract precise edge information. Additionally, \(f_1\) to \(f_4\) pass through the Selective Feature Decoupler (SFD) for feature map filtration. (3) The significant features outputted by the SFD are fed into the Decoder Blocks from higher to lower layers. The Edge Information Injector (EII) is utilized to interact the feature maps in the Decoder with edge information, thereby enhancing features related to edges. (4) Finally, the Feature Fusion Module (FFM) globally fuses the outputs of the multi-level Decoder Blocks and predicts the final Mask.

We will detail our proposed modules EME, EII, SFD, FFM, and the end-to-end design of our model in the following sections.

\subsubsection{Edge Mapping Engine.}

In our method, the Edge Mapping Engine (EME) plays a crucial role, designed to precisely extract the edge information of polyps from the feature maps outputted by the Encoder. This extracted information is then utilized to assist in the prediction of the polyp mask. 

Fig. \ref{fig_major}(b) illustrates the structure of the EME, which mainly consists of top-down and bottom-up stages. In the top-down stage, the feature map undergoes two downsampling processes, each stage comprising a CBR Block (Convolution, Batch Normalization, and ReLU), followed by max pooling. This process, while reducing spatial dimensions, effectively extracts the fundamental features of polyp edges.
Following the downsampling, the process enters the bottom-up stage. In this phase, the decoder part of the Edge Mapping Engine (EME) performs a series of bilinear interpolation upsampling operations. Subsequently, the features obtained from upsampling are fused along the channel direction with the corresponding features from the downsampling path. This fusion is vital for retaining the high-resolution details necessary for accurate edge prediction. 

After undergoing the aforementioned two stages, we obtain a feature map \(f_e\) enriched with precise edge information. This feature map is utilized in the Edge Information Injector (EII) to enhance the feature maps outputted by the Decoder. On the other hand, the \(f_e\) is also used for direct prediction of edges.

\begin{figure}[t]
\includegraphics[width=1\linewidth]{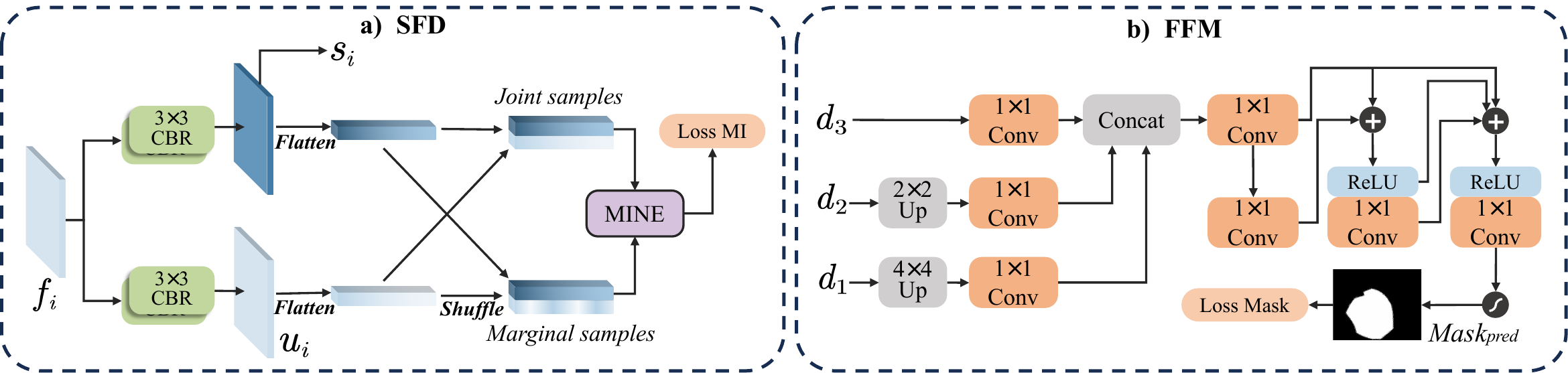}
\caption{a) Structure of the Selective Feature Decoupler (SFD). b) Structure of the Feature Fusion Module (FFM). } \label{fig_sfd_ffm}
\end{figure}

\vspace{-2mm}
\subsubsection{Edge Information Injector.}
Fig. \ref{fig_major}(c) shows the structure of the Decoder Block and the Edge Information Injector (EII). Our Decoder adopts the typical structure from \cite{tganet}. EII employs a serial dual attention mechanism, namely the Channel Attention Module (CAM) \cite{cbam} and Spatial Attention Module (SAM)\cite{cbam}, to process the feature map \(f_e\) provided by EME. Specifically, Ell first enhances the channels related to edge information in the feature map through the channel attention mechanism \(f_{ca}=CA\left(f_e\right)\). Then, it applies spatial attention \(f_{csa}=SA\left(f_{ca}\right)\) to further highlight the edge areas. The feature map outputted from the Decoder Block is concatenated with the feature map weighted by channel and spatial attention, and then passed through a \(1\times 1\) convolutional layer to produce the final feature map. In this way, EII further strengthens edge information both in the channel and spatial dimensions and injects the enhanced edge information into the Decoder, ensuring the model focuses more on the edge areas of the polyp in the segmentation task, thereby achieving a more accurate segmentation effect.

\vspace{-2mm}
\subsubsection{Selective Feature Decoupler.}
Due to the complexity and variability of the background in colonoscopy images, as well as varying lighting conditions, models are easily disturbed by a large number of irrelevant or unimportant features \cite{disturb}. 
We propose the Selective Feature Decoupler (SFD), a component bridging the Encoder and Decoder. By decoupling features and minimizing mutual information, it filters out ineffective information in the feature map. 

Fig. \ref{fig_sfd_ffm}(a) displays the structure of the Selective Feature Decoupler (SFD). Initially, the feature map \(f_i\) outputted from the \(i\)-th Encoder is processed through two separate sets of CBR blocks, resulting in the formation of \(s_i\), containing significant semantic information, and \(u_i\), primarily comprising unimportant information. Then, \(s_i\) is forwarded to the Decoder for subsequent processing, while \(u_i\) is discarded. A crucial aspect of the SFD's design is how it directs \(s_i\) and \(u_i\) to learn the semantic information we desire. To solve this, we employ an innovative Mutual Information Neural Estimation (MINE) \cite{mine} approach, which effectively estimates the lower bound of mutual information in high-dimensional data, allowing us to gauge the differences in semantic information between the two feature maps. 
This estimation method is based on the following definition of mutual information:
\begin{equation}
I(X ; Y)=\mathbb{E}_{P_{X Y}}\left[\log \frac{d P_{X Y}}{d\left(P_X \otimes P_Y\right)}\right],
\end{equation}
where \(\mathbb{E}\) represents the expectation. \(I(X ; Y)\) is the mutual information between the random variables \(X\) and \(Y\). \(P_{X Y}\) is their joint probability distribution, and \(P_X\) and \(P_Y\) are their respective marginal probability distributions.

To estimate mutual information, we form joint feature vectors by sequentially pairing flattened \(s_i\) with \(u_i\), and marginal feature vectors by randomly pairing them. Subsequently, these two vectors are fed into our constructed MINE network, a three-layer fully connected neural network. According to \cite{mine}, The MINE network outputs an estimated value of the mutual information, which can be expressed as follows:
\begin{equation}
\hat{I}(X; Y) = \mathbb{E}_{P_{XY}}[T_\theta(X, Y)] - \log\left(\mathbb{E}_{P_X \otimes P_Y}\left[e^{T_\theta(X, Y)}\right]\right),
\end{equation}
where  \(\hat{I}(X ; Y)\) is the estimated mutual information between the random variables \(X\) and \(Y\). \(P_{X Y}\) is their joint probability distribution, and \(P_X\) and \(P_Y\) are their respective marginal probability distributions. \(T_\theta\) is a function parameterized by the MINE network, used to estimate the logarithmic ratio between the joint and marginal distributions. 
Ultimately, we obtain \(loss_{MI}\) from 4 levels:
\begin{equation}
loss_{MI}=\sum_{i=1}^4{Sigmoid\left( \hat{I}\left( s_i;u_i \right) \right)},
\end{equation}
where, we employ the Sigmoid function to constrain the estimated value of mutual information within the range of 0 to 1. This enhances its numerical stability while preserving the monotonicity of the function, facilitating gradient backpropagation and optimization. 
Through the MINE network, the SFD is encouraged to learn to decompose the input feature map into significant and unimportant features, thereby effectively filtering out irrelevant information.

\vspace{-2mm}
\subsubsection{Feature Fusion Module.}

The Feature Fusion Module (FFM) captures rich contextual information and enhances the accuracy of polyp segmentation by integrating features from different decoder layers. Fig.\ref{fig_sfd_ffm}(b) clearly depicts the structure of the FFM. Initially, FFM aligns multi-scale feature maps from various decoder layers in space, using bilinear interpolation upsampling to ensure all feature maps have the same size, followed by a series of 1x1 convolution operations to standardize their channel counts. Subsequently, these feature maps are concatenated along the channel dimension and further integrated through another \(1\times1\) convolution layer. Finally, FFM enhances feature transmission through residual connections. By fully leveraging the complementarity of multi-scale features, FFM provides the model with a richer semantic information, thereby improving the accuracy and detail representation of the final segmentation results. 


\vspace{-2mm}
\subsubsection{End-to-End Design.}
It is noteworthy that our EPPS is designed to be fully end-to-end. On one hand, the MINE network within the SFD is embedded in the entire model and does not require any pre-training. On the other hand, the ground truth of polyp edges do not require manual annotation, offering maximum convenience for the model's training.
Specifically, during the training of the model, we utilize edge detection operators (such as the Canny or Sobel operators) to extract polyp edges from the ground truth of polyp masks. These extracted edges serve as the ground truth for edge prediction by the EME. Compared to manually annotating edge images, our method is convenient and precise, completely avoiding uncertainties brought about by manual operations, ensuring that the ground truth of the edges perfectly match the ground truth mask of polyps.
For the edge predictions outputted by the EME and the final polyp segmentation mask outputted by EPPS, we calculate their respective loss functions:
\begin{equation}
    loss_{mask} = L(mask_{pred}, mask_{gt}),~loss_{edge} = L(edge_{pred}, edge_{gt}),
\end{equation}
where \(mask_{pred}\) and \(mask_{gt}\) respectively represent the polyp mask predicted by EPPS and its ground truth, while \(edge_{pred}\) and \(edge_{gt}\) correspond to the edge predicted by EME and the edge's ground truth. \(L(\cdot)\) denotes the DiceBCE loss, which is obtained by summing the Dice loss and the BCE loss.
Additionally, the mutual information loss output by the SFD is incorporated into the final joint loss function, which can be represented as: 
\begin{equation}
    loss_{joint} = loss_{mask} + \alpha~loss_{edge} + \beta~loss_{MI},
\end{equation}
where \( \alpha \) and \( \beta \) are weight parameters used to balance these losses.

\begin{figure}[t]
    \centering
    \includegraphics[width=1\linewidth]{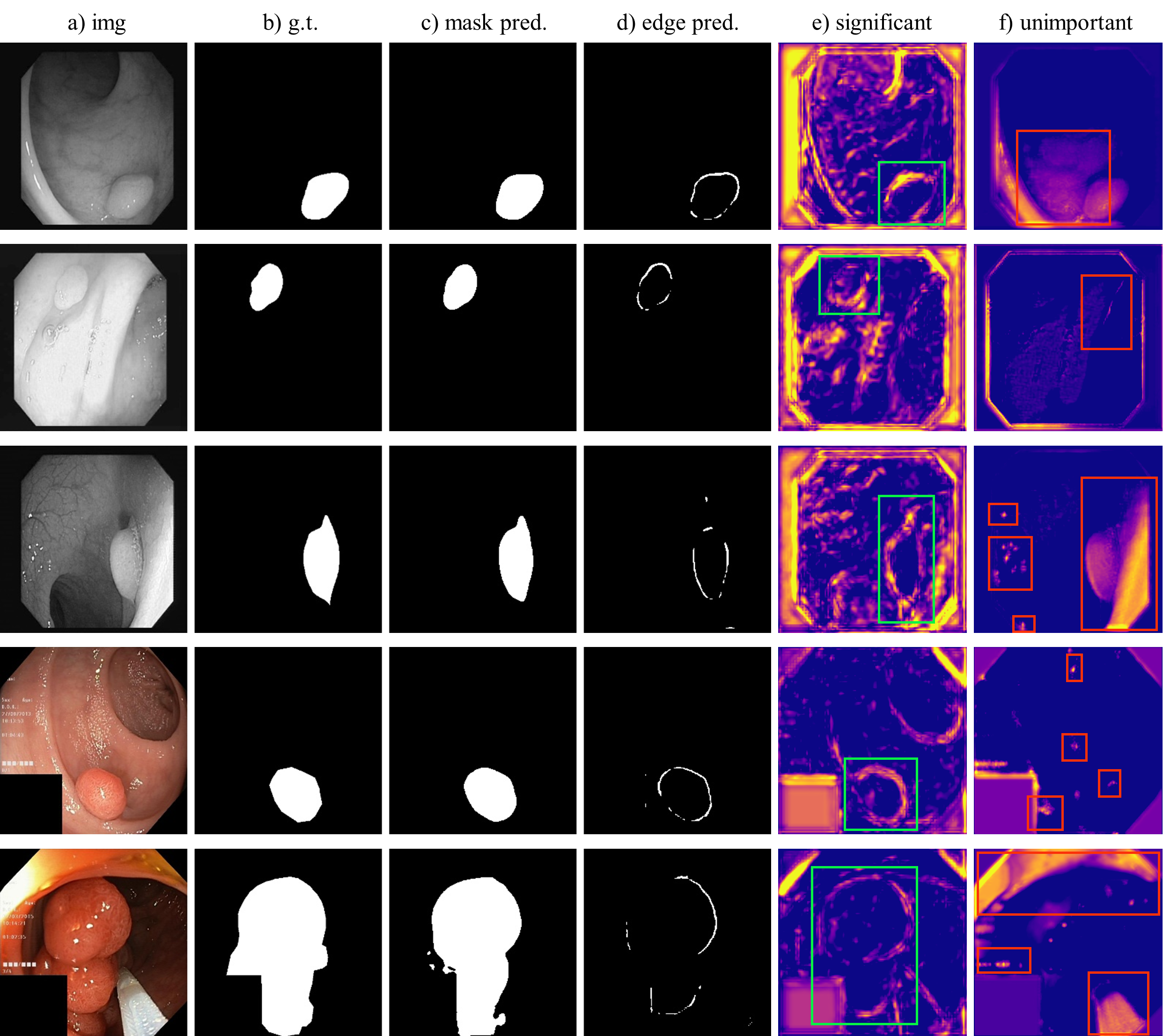}
    \caption{Additional results on CVC-ClinicDB and Kvasir-SEG datasets. In e), for the significant feature maps (\(s_i\)), we outline the edge-related features learned by SFD with green boxes. In f), for the unimportant features (\(u_i\)), we outline the bright areas with red boxes, primarily located in areas of reflection and foreign objects, which are ineffective or could even mislead the model. }
    \label{fig_vis}
\end{figure}

\section{Experiments and Results}



\subsubsection{Datasets and Implementation Details.}
To evaluate the performance of EPPS, we utilized three publicly available polyp segmentation benchmark datasets, including Kvasir-SEG\cite{kvasir_seg}, CVC-ClinicDB\cite{cvc_clinicdb}, and Kvasir-Sessile\cite{kvasir_seg}. The CVC-ClinicDB comprises 612 polyp images collected from colonoscopy examination video sequences. Kvasir-SEG includes 1,000 polyp images sourced from multiple colonoscopy examination video sequences. Kvasir-Sessile is a smaller dataset, containing 196 images of small sessile and flat polyps, each less than 10 millimeters in size.
All experiments are conducted on an NVIDIA GeForce RTX 3090 GPU, with image sizes adjusted to \(256\times256\) pixels. We adopt an \(8:1:1\) split for training, validation, and testing. 
A simple data augmentation strategy is employed, including random rotation, vertical flipping, and horizontal flipping. 
For the joint loss function, we set both alpha and beta to a default value of \(1\). The learning rate is set to \( 1e^{-4}\), and Adam optimizer and early stopping mechanism are utilized while training. We employ 4 popular evaluation metrics from prior studies \cite{tganet,ldnet}: the mean Dice Similarity Coefficient (mDSC), mean Intersection over Union (mIoU), Recall,  Precision.

\begin{table}[t]
\centering
\caption{Comparison with other state-of-the-art methods on Kvasir-SEG and CVC-ClinicDB dataset.}\label{tab_seg_clinicdb}
\begin{tabularx}{\textwidth}{@{}l*{8}{X}@{}}
\toprule
\multirow{2}{*}{Method} & \multicolumn{4}{c}{Kvasir-SEG} & \multicolumn{4}{c}{CVC-ClinicDB} \\ \cmidrule(l){2-5} \cmidrule(l){6-9} 
& \(mIoU\) & \(mDSC\) & \(Rec\) & \(Prec\) & \(mIoU\) & \(mDSC\) & \(Rec\) & \(Prec\) \\ \midrule
U-Net\cite{unet}       & 0.747 & 0.826 & 0.850 & 0.870 & 0.843 & 0.898 & 0.900 & 0.921 \\
PraNet\cite{pranet}      & 0.830 & 0.894 & 0.906 & 0.913 & 0.887 & 0.932 & 0.935 & 0.948 \\
ColonSeg\cite{colonseg} & 0.698 & 0.792 & 0.819 & 0.873 & 0.824 & 0.886 & 0.883 & 0.902 \\
TGANet\cite{tganet}      & 0.833 & 0.898 & 0.913 & 0.912 & \underline{0.899} & \underline{0.946} & 0.944 & \underline{0.952} \\
MSRF-Net\cite{msrf}~    & 0.740 & 0.851 & 0.807 & 0.899 & 0.828 & 0.906 & 0.862 & \textbf{0.955} \\
LDNet\cite{ldnet}       & \underline{0.853} & \underline{0.907} & \underline{0.927} & \underline{0.920} & 0.895 & 0.943 & \underline{0.945} & 0.945 \\
\textbf{EPPS(Ours)}  & \textbf{0.876} & \textbf{0.930} & \textbf{0.933} & \textbf{0.939} & \textbf{0.904} & \textbf{0.948 }& \textbf{0.946} & \textbf{0.955} \\ 
\bottomrule
\end{tabularx}
\end{table}

\begin{table}[t]
\centering
\caption{Comparison with other state-of-the-art methods on Kvasir-Sessile dataset.}\label{tab_compare_kvasir_sessile}
\begin{tabular}{lccccccc}
\toprule
Metrics & UNet\cite{unet}& PraNet\cite{pranet} &ConlonSeg\cite{colonseg}& TGANet\cite{tganet} & XBound\cite{xbound} &DTAN\cite{dtan} & \textbf{Ours}\\
\midrule
\(mDSC\)~ & 0.369& 0.774 &0.328& 0.820& 0.811&\underline{0.842}& \textbf{0.874}\\
\(mIoU\)~& 0.247& 0.667 &0.211& 0.744& 0.736&\underline{0.764}& \textbf{0.784}\\
 \(Rec\)& 0.724& 0.807 &0.523& 0.793& \textbf{0.872}& 0.842&\underline{0.862}\\
 \(Prec\)& 0.326& 0.824 &0.334& \underline{0.859}& 0.763& \underline{0.859}&\textbf{0.914}\\
 \bottomrule
\end{tabular}
\end{table}

\vspace{-2mm}
\subsubsection{Comparison with Other Methods.}
We compared EPPS with other advanced methods, as shown in Table \ref{tab_seg_clinicdb} and Table \ref{tab_compare_kvasir_sessile}. It achieved the best performance on Kvasir-SEG, CVC-ClinicDB, and Kvasir-Sessile datasets.
On Kvasir-SEG, our method achieved the best performance across all metrics. Notably, EPPS showed an improvement of \(2.3\%\) in both mIoU and mDSC compared to the second-best performing method. On CVC-ClinicDB, EPPS continued to outperform all others. On Kvasir-Sessile, EPPS exceeded the second-best models by \(3.2\%\), \(2.0\%\), and \(4.5\%\) in mIoU, mDSC, and Precision, respectively. Overall, our method achieved state-of-the-art performance on datasets of various sizes, demonstrating its efficiency and robustness.

\begin{table}[t]
\centering
\begin{minipage}{.48\textwidth}
  \centering
  \caption{Impact of using different components on model performance.}\label{tab_ablation}
  \begin{tabular}{ccccc}
    \toprule
    Methods & \(mDSC\) & \(mIoU\) & \(Rec\) & \(Prec\) \\
    \midrule
    Baseline & 0.905 & 0.853 & 0.912 & 0.920 \\
    +SFD & 0.927 & 0.882 & 0.922 & 0.940 \\
    +EME\&EII & 0.941 & 0.893 & 0.944 & 0.943 \\
    +All & 0.948 & 0.904 & 0.946 & 0.955 \\
    \bottomrule
  \end{tabular}
\end{minipage}%
\hfill 
\begin{minipage}{.49\textwidth}
  \centering
  \caption{Impact of setting different \(\alpha\) and \(\beta\) values on model performance.}\label{tab_alpha_beta}
  \begin{tabular}{cccccccc}
    \toprule
    \multirow{2}{*}{Metrics} & \multicolumn{3}{c}{\(\alpha\)} & \multicolumn{3}{c}{\(\beta\)} \\
    \cmidrule(lr){2-4} \cmidrule(lr){5-7}
    & 1 & 0.1 & 0.01 & 1 & 0.1 & 0.01 \\
    \midrule
    mIoU & 0.893 & 0.882 & 0.880 & 0.882 & 0.879 & 0.871 \\
    mDSC & 0.941 & 0.930 & 0.927 & 0.927 & 0.926 & 0.918 \\
    \bottomrule
  \end{tabular}
\end{minipage}
\end{table}

\begin{table}[t]
\centering
\caption{Impact of choosing different edge detection operators on model performance.}\label{tab_different_edge_det}
\begin{tabularx}{0.8\textwidth}{l*{5}{>{\centering\arraybackslash}X}}
\toprule
Metrics~ & Sobel & Laplacian & Canny & Scharr & Prewitt \\
\midrule
\(mIoU~\) & 0.855 & 0.871 & 0.876 & 0.860 & 0.851 \\
\(mDSC~\) & 0.913 & 0.921 & 0.930 & 0.919 & 0.911 \\
\bottomrule
\end{tabularx}
\end{table}

\subsubsection{Ablation Study.}

\begin{figure}[h]
    \centering
    \includegraphics[width=1\linewidth]{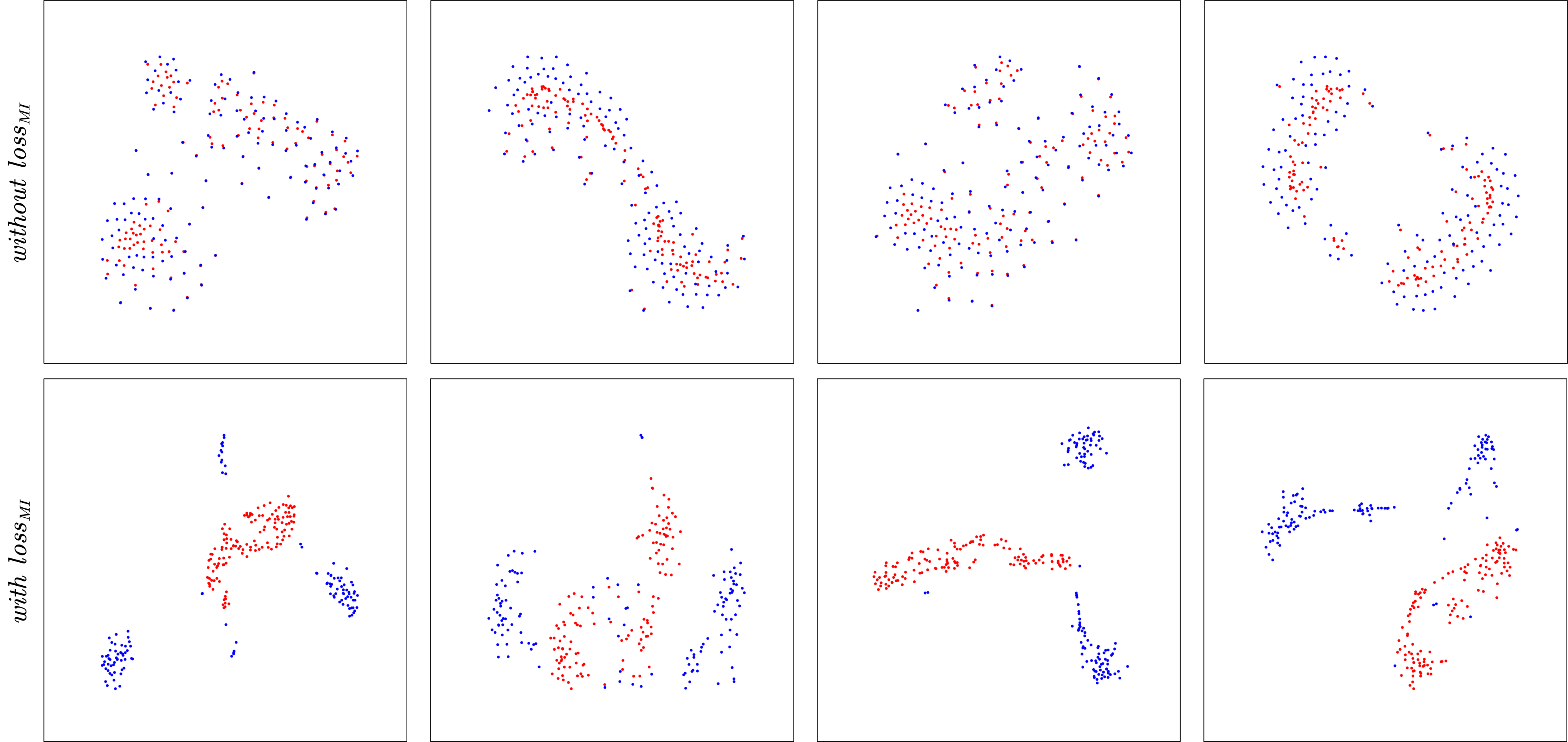}
    \caption{The t-SNE visualization of the two features outputted by SFD (\(s_i\) and \(u_i\)) with and without mutual information constraint. Red indicates \(s_i\), while blue indicates \(u_i\). }
    \label{fig:tsne}
\end{figure}

\begin{figure}[h]
    \centering
    \includegraphics[width=0.5\linewidth]{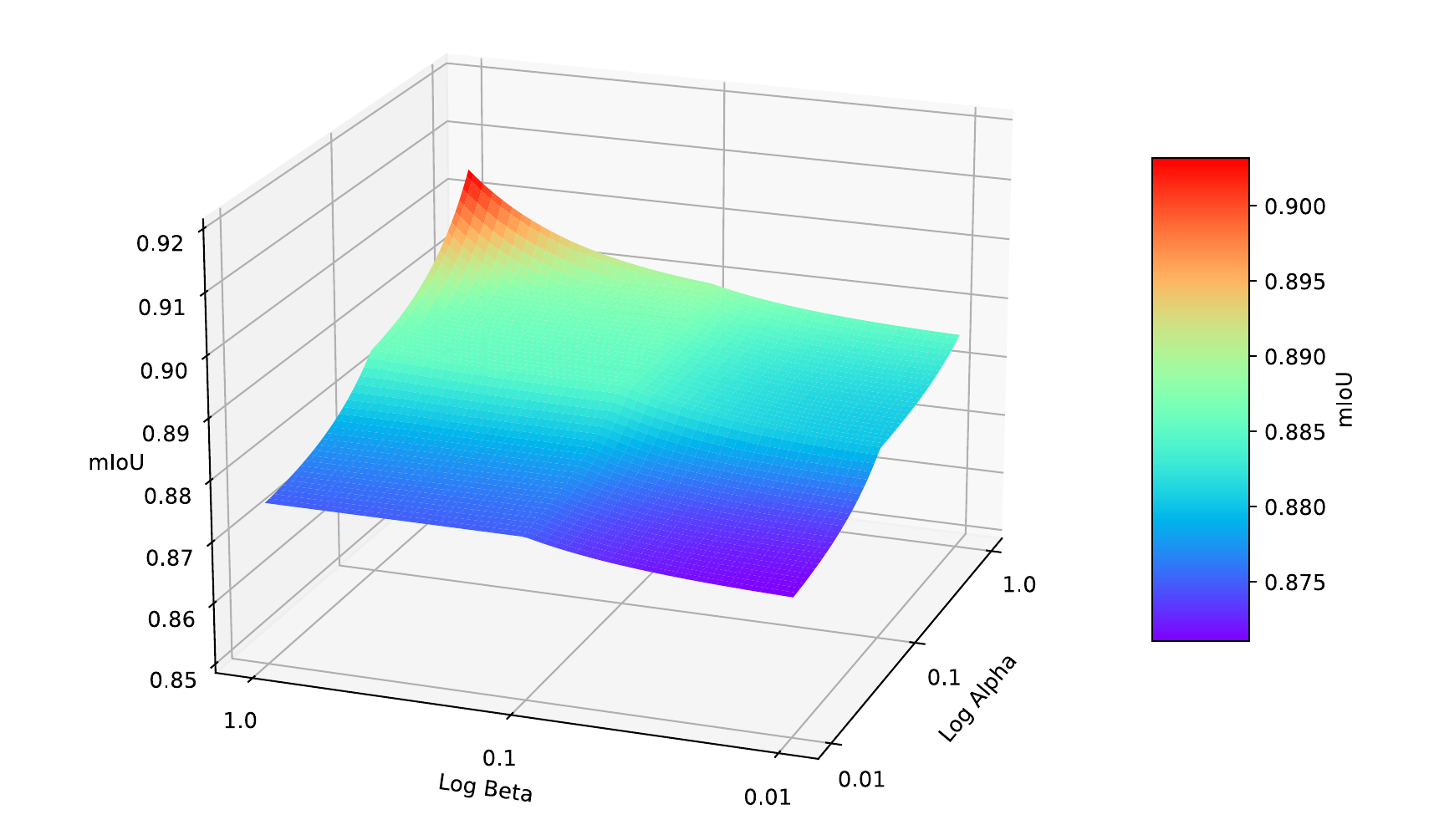}
    \caption{Visualization of the model's mIoU scores on the CVC-ClinicDB dataset under different combinations of \(\alpha\) and \(\beta\), using a 3D surface plot. }
    \label{fig:surface}
\end{figure}

To validate the effectiveness of our proposed core components, we conducted four sets of ablation experiments. Specifically, we removed the EME and EII modules and replaced the SFD with a standard \(3×3\) convolution block as the Baseline. We evaluated the performance of the model under four different scenarios on the CVC-ClinicDB dataset: 1) baseline, 2) adding SFD alone, 3) adding EME and EII alone, 4) adding all modules together. Here, we set both \(\alpha\) and \(\beta\) values in the loss function to \(1\) and chose the Canny operator for extracting the ground truth of polyp edges. Table \ref{tab_ablation} displays the results of these experiments. Compared to the baseline, the addition of SFD and the addition of EME and EII both significantly improved model performance. When all components were added simultaneously, the model achieved the best performance in all metrics. Notably, mDSC and mIoU improved by \(4.3\%\) and \(5.1\%\), respectively, which are very significant improvements. 

Fig. \ref{fig_vis} demonstrates the output results of EPPS. As seen from d), EME accurately extracts the edge regions of the polyp, which plays a crucial role in aiding the prediction of the polyp mask. Additionally, as shown in e) and f), the significant feature map decoupled by SFD outlines the edges of the intestinal tissue, with the highlighted areas focusing on the edges of the polyp. In contrast, the highlighted areas of the unimportant features are more uniformly distributed within the interior rather than the edges of the intestinal tissue and are also likely to concentrate in the bright parts of the image, which are not key to the accurate segmentation of polyps or even misleading to the model's judgement. These are considered either unimportant or potentially misleading features. 
Fig. \ref{fig:tsne} further demonstrates the role of mutual information constraints in the SFD feature decoupling process. It shows the visualization using t-SNE of the two features (significant feature and unimportant feature) output by SFD, with and without mutual information constraints. We present four examples, from which it can be seen that the distributions of the two features output by SFD without mutual information constraints are similar. However, the outputs from SFD with mutual information constraints show significant differences in their distributions, indicating that our designed MINE network and \(loss_{mi}\) can effectively guide the feature decoupling.

To investigate the impact of varying the weight coefficients, \(\alpha\) and \(\beta\), for \(loss_{edge}\) and \(loss_{MI}\) in the loss function on performance, we altered the parameters for experiments 2) and 3) mentioned above and observed the model's performance. As shown in Table \ref{tab_alpha_beta}, the best model performance occurred when both alpha and beta were set to \(1\). To further explore the effect of different combinations of \(\alpha\) and \(\beta\) on model performance in the full EPPS, we conducted a number of experiments, visualised in Fig. \ref{fig:surface}.

Lastly, we investigated the impact of different edge extraction operators on our model using the Kvasir-SEG dataset. As shown in Table \ref{tab_different_edge_det}, we tested five operators: Sobel, Laplacian, Canny, Scharr, and Prewitt. Among these, the best results were achieved when utilizing the Canny operator.

\vspace{-1mm}
\section{Conclusion}
We proposed a novel polyp segmentation method, EPPS, specifically designed to address the challenges of discerning polyp edges and susceptibility to ineffective feature interference prevalent in existing methods. 
Utilizing the EME for precise edge extraction and the EII to incorporate edge information into Decoder Blocks, our method improves polyp mask prediction. Furthermore, we developed the SFD to suppress the influence of noise and extraneous features on the model. Our extensive experiments across three widely used datasets confirmed EPPS's state-of-the-art performance, underscoring its potential in enhancing clinical colonoscopy outcomes.

%
%
%
\bibliographystyle{splncs04}
\bibliography{mybibliography}

\end{document}